\documentclass[runningheads]{llncs}
\usepackage{amsmath}
\usepackage{url}
\usepackage[ruled,vlined]{algorithm2e}
\usepackage{float}
\usepackage{makecell}
\usepackage{amssymb}
\usepackage[normalem]{ulem}
\usepackage{placeins}
\usepackage[T1]{fontenc}
\usepackage{graphicx}
\begin{document}
\title{Bloom Filter Encoding for Machine Learning}
\author{John Cartmell\orcidID{0000-0002-7014-4005}%
\and Mihaela Cardei$^{\star}$\orcidID{0000-0003-2359-6196}%
\and Ionut Cardei
\thanks{These authors contributed equally to this work.}\orcidID{0009-0000-1050-768X}%
\thanks{Accepted for publication in the Proceedings of the 22nd Artificial Intelligence Applications and Innovations Conference (AIAI 2026), IFIP AICT Series, Springer.}
}
\authorrunning{J. Cartmell et al.}

\institute{Department of Electrical Engineering and Computer Science\\ Florida Atlantic University\\ Boca Raton FL 33431, USA\\ 
\email{\{jcartmell2023,mcardei,icardei\}@fau.edu}}
\maketitle
\begin{abstract}
We present a method that uses a Bloom filter transform to preprocess data for machine learning. Each sample is encoded into a compact bit-array representation using hash-based encoding, producing a fixed-length feature space that reduces memory usage and obfuscates original feature values. The encoding does not rely on keyed hashing; however, a key can optionally be used to control the mapping and would be required to reproduce the representation. We evaluate the approach on six datasets spanning text, time-series, tabular, and image domains: SMS Spam Collection, ECG200, Adult 50K, CDC Diabetes, MNIST, and Fashion MNIST. Four classifiers are considered: Extreme Gradient Boosting, Deep Neural Networks, Convolutional Neural Networks, and Logistic Regression. Results show that models trained on Bloom filter encodings achieve performance comparable to models trained on raw data or standard dimensionality reduction techniques across several datasets, while providing consistent memory savings. These findings suggest that Bloom filter encodings can serve as an efficient, general-purpose preprocessing representation that preserves useful similarity structure for learning tasks while providing a degree of data obfuscation.
\end{abstract}
\section{Introduction}
General-purpose preprocessing methods for machine learning that are both memory\allowbreak-efficient and privacy-aware remain limited. Most existing approaches optimize either compactness or privacy, and rarely achieve both in a unified representation \cite{NayakPatgiri2021}, \cite{Vidanage2020}. A practical method that reduces memory while obfuscating sensitive feature values, without substantially degrading predictive performance, is therefore of interest.

Bloom filters have long been studied as space-efficient probabilistic data structures for set membership testing \cite{Bloom1970}. In machine learning, they have been used as auxiliary structures before training or after inference to improve efficiency or support privacy-sensitive applications \cite{Mitzenmacher2002}, \cite{Broder2005}, \cite{Nitz2023}, \cite{Randall2022}. Related work has also explored hashing-based representations, such as feature hashing \cite{Weinberger2009}, and Bloom filter encodings for similarity-based tasks \cite{Nitz2023}. However, these approaches are typically applied in specific domains or as components within larger pipelines, rather than as general-purpose feature representations used directly for model training across diverse data modalities.

In this work, we introduce a method that encodes raw sample data using a Bloom filter transform. Each sample is mapped to a fixed-length binary representation via hash-based encoding, producing compact encodings suitable for direct use in machine learning models. The method does not rely on keyed hashing; however, a key can optionally be used to control the mapping and would be required to reproduce the representation. Unlike prior work that applies Bloom filters within specific tasks or as auxiliary structures, this work uses Bloom filter encodings as a general-purpose preprocessing representation applied directly to raw data across multiple modalities and models. This representation reduces memory usage \cite{NayakPatgiri2021} and provides a degree of data obfuscation through hashing and collisions \cite{Vidanage2020}, while empirically retaining sufficient structure for accurate learning.

We evaluate the transform on six publicly available datasets spanning text, time-series, tabular, and image types: SMS Spam Collection \cite{UCI_SMS2011}, ECG200 \cite{Olszewski2001}, Adult 50K \cite{DuaGraff1996}, CDC Diabetes \cite{CDC2020}, MNIST \cite{LeCun1998}, and Fashion MNIST \cite{FashionMNIST2017}. We use four classifiers: Extreme Gradient Boosting (XGB) \cite{Chen2016}, Deep Neural Networks (DNN) \cite{Hinton2006}, Convolutional Neural Networks (CNN) \cite{Krizhevsky2012}, and Logistic Regression (LR) \cite{Cox1958}. 

Results show that the Bloom filter transform achieves classification performance comparable to raw data across several tabular and time-series datasets, with improvements in some cases. For example, on the EKG dataset, accuracy increases from 81.0\% (raw) to 82.9\% (+1.9\%), and on Adult 50K using a DNN classifier, accuracy rises from 88.1\% to 88.9\% (+0.8\%). On image datasets such as MNIST and Fashion MNIST, accuracy decreases from 98.1\% to 95.1\% (-3.0\%) and from 90.5\% to 85.3\% (-5.2\%), respectively, likely due to the loss of spatial structure when mapping images into hashed representations. In addition to maintaining predictive performance, the Bloom filter transform provides compression, reducing memory usage by approximately 2--4$\times$ compared to raw data. Entropy (0.38--0.68) and bit occupancy (0.13--0.60) indicate that the representation balances information density and collision effects, contributing to both efficiency and data obfuscation. Parameter sweeps further show that filter size and number of hash functions can be tuned to balance accuracy and compression, enabling flexible trade-offs for different applications.

The rest of the paper is organized as follows. Section 2 presents Bloom filter design, metrics, and characteristics. Section 3 describes transforms and encoding methods. Section 4 details the pipeline and its application to multiple datasets and classifiers. Section 5 presents results and analysis. Section 6 outlines future directions.
\section{Bloom Filters}
\begin{figure}[t]
    \vspace{-2mm}
    \centering
    \includegraphics[width=.9\textwidth]{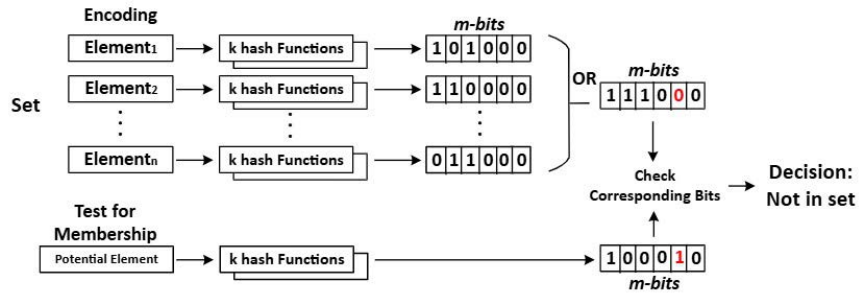}
    \caption{Bloom filter construction/test process for element not a member of set.}
    \label{fig:bf_construct}
    \vspace{-4mm}
\end{figure}

Bloom filters \cite{Bloom1970} are compact probabilistic data structures used to test set membership efficiently. Each element is hashed by $k$ independent functions, setting bits in an $m$-bit array. The final Bloom filter is obtained by combining all insertions through a logical OR operation, as illustrated in Fig. \ref{fig:bf_construct}. To test membership, a candidate element is hashed and the corresponding bits are checked: if any bit is unset, the element is absent; if all are set, it may be present.

\subsection{Construction and Design Considerations}
Designing a Bloom filter involves balancing memory usage, collision effects, and false positive behavior. Each inserted element sets $k$ bits in the $m$-bit array. The expected fraction of bits set (bit occupancy) after inserting $n$ elements is \cite{Bloom1970}:
\begin{equation}
    p_1 = 1 - e^{-kn/m} \label{eq:bit_occupancy}
\end{equation}

As occupancy increases, collisions become more likely, which in turn increases the false positive rate (FPR), given by \cite{Broder2005}:
\begin{equation}
    \text{FPR} = (1 - e^{-kn/m})^k = p_1^k \label{eq:fpr}
\end{equation}

These relationships highlight the trade-off between representation density and collision effects. Smaller filters yield higher compression but increased collisions, while larger filters reduce collisions at the cost of additional memory.

The compression ratio is defined as \cite{Mitzenmacher2002}:
\begin{equation}
    \text{CR} = \frac{n \cdot S}{m} \label{eq:compression}
\end{equation}
where $S$ is the size of each element in bits. Bloom filters provide fixed-length representations independent of the number of inserted elements, making them attractive for compact data encoding.

Entropy provides a measure of information density in the representation. For a bit with activation probability $p_1$, the entropy is:
\begin{equation}
    H_\text{bit} = -\big[p_1 \log_2 p_1 + (1-p_1) \log_2 (1-p_1)\big] \label{eq:entropy}
\end{equation}
Entropy is maximized near $p_1 = 0.5$, while very low or high occupancy corresponds to low information content.

\subsection{Hashing and Reconstruction Considerations}
The behavior of Bloom filter encodings depends on the choice of hash functions, the number of hashes $k$, and the resulting bit occupancy. As occupancy increases, multiple elements map to overlapping bit positions, making it increasingly difficult to associate individual bits with specific input features.

Standard hash functions are sufficient for constructing Bloom filter encodings. Keyed cryptographic hash functions (e.g., HMAC-SHA256) can optionally be used to further obscure the mapping between input features and bit positions \cite{Krawczyk1997}. While such constructions can make direct reconstruction of inputs more difficult, they do not provide formal privacy guarantees on their own. Instead, they contribute to a degree of data obfuscation by distributing feature information across the representation and introducing ambiguity through collisions.
\section{Transforms}
A transform is a mathematical operation that maps data from one representation into another according to a defined rule \cite{Liu2022}. In machine learning, transforms are commonly used to improve representation efficiency, highlight relevant structure, or reduce dimensionality. Bloom filters, while traditionally used for set membership, can also be interpreted as a transform that maps data from the original feature space into a fixed-length bit-array representation.

Table \ref{tab:distance_transforms} summarizes several commonly used transforms and their approximate effects on distance or structure.

\begin{table}[h]
\centering
\vspace{-4mm}
\caption{Representative transforms and their structural properties}
\label{tab:distance_transforms}
\begin{tabular}{|l|p{7cm}|}
\hline
\textbf{Transform} & \textbf{Structural Properties} \\ \hline
Identity \( T(x) = x \) & Perfect preservation \\ \hline
Scaling \( T(x) = a x \) & Distances scaled uniformly \\ \hline
Rotation / Orthogonal transform & Preserves Euclidean distances \\ \hline
PCA & Preserves dominant variance directions \\ \hline
LDA & Preserves class-discriminative structure \\ \hline
Bloom filter encoding & Preserves approximate similarity structure via hashing \\ 
\hline
\end{tabular}
\end{table}

Transforms that preserve useful structure enable downstream models to learn meaningful decision boundaries. In contrast, transforms that destroy structure (e.g., random shuffling) degrade model performance. Unlike linear transforms such as PCA or LDA, Bloom filter encodings are non-linear and rely on hashing, which introduces collisions but also enables fixed-length representations independent of input dimensionality.

Both Principal Component Analysis (PCA) \cite{jolliffe2002} and Linear Discriminant Analysis (LDA) \cite{fisher1936} reduce dimensionality by projecting data into a lower-dimensional space. PCA maximizes variance, while LDA emphasizes class separability \cite{anowar2021}. In contrast, Bloom filter encodings do not explicitly optimize for variance or class separation; instead, they provide a compact representation that distributes feature information across a shared bit space. This distinguishes Bloom filter encodings from traditional dimensionality reduction methods, which aim to preserve variance or class separability, whereas the proposed approach emphasizes compactness and approximate similarity preservation.

\subsection{Bloom Filter Transform}
Bloom filter encodings can be interpreted as preserving approximate similarity structure rather than exact distances. Consider two data points $x, y \in \mathbb{R}^d$ (or sets of features) and their corresponding Bloom filter encodings $b(x), b(y) \in \{0,1\}^m$. The Hamming distance between these encodings is:
\begin{equation}
d_H(b(x), b(y)) = \sum_{i=1}^{m} \mathbf{1}\{ b(x)_i \neq b(y)_i \},
\end{equation}
where $\mathbf{1}\{\cdot\}$ is the indicator function.

Because hashing is deterministic, identical feature values map to consistent bit positions, and shared features contribute to overlapping bit patterns. As a result, inputs with greater feature overlap tend to produce more similar Bloom filter encodings. However, due to collisions and the non-linear nature of hashing, this relationship is approximate rather than exact.

In expectation over multiple hash functions, the Hamming distance between Bloom filter encodings can reflect relative similarity between inputs \cite{Broder2005}, \cite{Nitz2023}, \cite{Vidanage2020}:
\begin{equation}
\mathbb{E}[d_H(b(x), b(y))] \approx F \,( D_{\text{orig}}(x, y)),
\end{equation}
where $D_{\text{orig}}(x, y)$ is a distance metric in the original space and $F$ is a monotonic function induced by the hashing and tokenization process. The exact form of $F$ depends on the choice of hash functions and quantization and is not explicitly defined. This implies that relatively closer points in the original space tend to remain closer in the encoded space, although this relationship is not guaranteed for all pairs.

This similarity preservation depends on the tokenization step, where continuous values are quantized prior to hashing, ensuring that similar inputs produce overlapping feature tokens. Each feature-value pair is represented as a token of the form (feature name, quantized value). This approximate preservation of similarity is sufficient for many machine learning tasks, where exact distance preservation is not required. Instead, models rely on consistent patterns in the feature space to learn decision boundaries. Bloom filter encodings provide such patterns while enabling compression and obfuscation through hashing and collisions.

While Bloom filter transforms differ fundamentally from classical distance-preserving transforms, their ability to retain useful similarity structure in practice makes them a viable preprocessing approach. This behavior is evaluated empirically in the experimental sections that follow.
\section{Methodology}
\begin{figure}[tb]
    \vspace{-2mm}
    \centering
    \includegraphics[width=.95\textwidth]{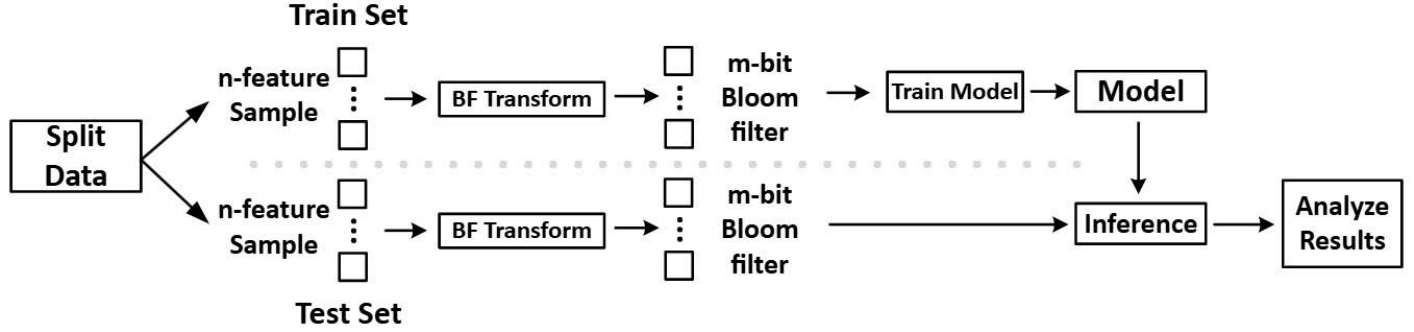}
    \caption{Training/test pipeline: feature encoding into $m$-bit Bloom filter representations, model training, inference, and evaluation.}
    \label{fig:generic_pipeline}
    \vspace{-4mm}
\end{figure}

The overall workflow is shown in Fig. \ref{fig:generic_pipeline}. The process begins with data collection, followed by a training/test split. Each sample is then transformed into a Bloom filter bit array. Bloom filter encodings derived from the training set are used to train models, while those from the test set are used for evaluation. Model predictions are compared against ground truth labels. In addition to predictive performance, compression and representation characteristics of the Bloom filter encodings are analyzed.

\subsection{Datasets}
Six publicly available datasets were used, spanning text, time-series, tabular, and image data types. The SMS Spam Collection dataset contains 5,574 English messages labeled as `ham' or `spam' \cite{UCI_SMS2011}. The EKG (ECG200) dataset consists of 200 heartbeat recordings, evenly split between normal and abnormal rhythms \cite{Olszewski2001}. Tabular datasets include Adult 50K (48,842 samples, 14 features, income classification) \cite{DuaGraff1996} and CDC Diabetes (250,000 samples, 23 features) \cite{CDC2020}. Image datasets comprise MNIST (70,000 handwritten digits across 10 classes) \cite{LeCun1998} and Fashion MNIST (70,000 fashion images across 10 classes) \cite{FashionMNIST2017}. For datasets without predefined splits, 5-fold cross-validation was used.

\subsection{Encoded Bloom Filter Transform}
\begin{figure}[tb]
    \vspace{-2mm}
    \centering
    \includegraphics[width=.95\textwidth]{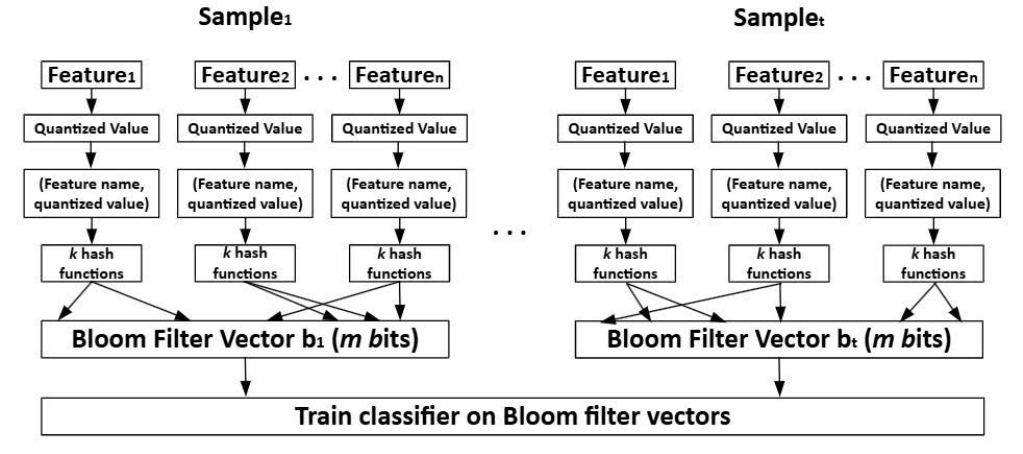}
    \caption{Transformation of samples into Bloom filter bit arrays. Continuous feature values are quantized into tokens prior to hashing, enabling approximate similarity preservation in the encoded space.}
    \label{fig:transform}
    \vspace{-4mm}
\end{figure}

Each sample is transformed into an $m$-bit Bloom filter vector, which serves as the input representation for downstream models (Fig. \ref{fig:transform}, Algorithm \ref{alg:generic_preprocessing}). Each row of $\mathbf{B}$ corresponds to the Bloom filter vector for one sample.

Prior to encoding, continuous feature values are quantized into discrete representations. Each feature is represented as a token of the form (feature name, quantized value). This enables similar inputs to map to shared tokens, supporting approximate similarity preservation in the encoded space. Each token is inserted into the Bloom filter using $k$ hash functions.

The resulting representation is a fixed-length binary vector $\mathbf{b} \in \{0,1\}^m$, independent of the original feature dimensionality. The bit array may be stored in a byte-aligned format; however, the logical size remains $m$.

A family of $k$ deterministic hash functions $h_1, \dots, h_k$ maps tokens to indices in $[0, m-1]$:
\begin{equation}
h_i(f, v) = H(f \,\|\, v \,\|\, i) \bmod m, \quad i = 1, \dots, k
\label{eq:hash_family}
\end{equation}
where $H(\cdot)$ is a deterministic hash function and $\|$ denotes concatenation. In this work, HMAC-SHA256 \cite{NIST2024} is used as the underlying hash function; the use of a key is optional and only required for reproducibility.

\begin{algorithm}[H]
\caption{Bloom Filter Encoding of Samples}
\label{alg:generic_preprocessing}
\KwIn{Samples $\{s_1,\dots,s_N\}$, Bloom filter size $m$, number of hash functions $k$}
\KwOut{Bloom filter representation matrix $\mathbf{B} \in \{0,1\}^{N \times m}$}

Initialize $\mathbf{B}$ as an $N \times m$ binary matrix set to 0\;

\For{$j \gets 1$ \KwTo $N$}{
    Initialize $\mathbf{b}_j$ as an $m$-bit vector set to 0\;
    \ForEach{feature $f$, value $v$ in $s_j$}{
        $v_q \gets \text{Quantize}(v)$\;
        \For{$i \gets 1$ \KwTo $k$}{
            $idx \gets h_i(f, v_q)$\;
            $\mathbf{b}_j[idx] \gets 1$\;
        }
    }
    $\mathbf{B}[j,:] \gets \mathbf{b}_j$\;
}
\Return $\mathbf{B}$\;
\end{algorithm}

\subsection{Model Training and Inference}
Each training sample is transformed into its Bloom filter representation and used as input to classification models. We evaluate four classifiers: Logistic Regression (LR) \cite{Cox1958}, Extreme Gradient Boosting (XGB) \cite{Chen2016}, Deep Neural Networks (DNN) \cite{Hinton2006}, and Convolutional Neural Networks (CNN) \cite{Krizhevsky2012}. Test samples undergo the same transformation prior to inference.

Using multiple model classes allows us to evaluate whether the Bloom filter representation retains sufficient structure for learning across different modeling paradigms. In particular, tree-based and linear models operate directly on the binary representation, while neural models assess whether more complex patterns can be learned from the encoded space. CNN performance also provides insight into how well the representation preserves local structure, which is not explicitly maintained under hashing.

\subsection{Analysis}
We evaluate the following metrics:
\begin{itemize}
    \item Classification performance (Accuracy, $F_1$, AUC)
    \item Compression Ratio 
    \item Entropy and Bit Occupancy
\end{itemize}

For each dataset, Bloom filter parameters ($m$, $k$) are varied to analyze trade-offs between predictive performance and representation size. Results are compared against baseline representations, including raw features and, where applicable, PCA and LDA encodings. For PCA and LDA, the number of components was selected such that the resulting representation size was approximately matched to the Bloom filter encoding, enabling a fair comparison of predictive performance under similar memory constraints.

$F_1$ of the positive class is used for SMS Spam and EKG datasets, where minority class detection is important \cite{Vidanage2020}. For Adult 50K and CDC Diabetes, AUC is used due to class imbalance \cite{Chen2016}. For image datasets, macro $F_1$ is reported across classes \cite{Randall2022}.

\begin{table}
\caption{Datasets, Encoders, Classifiers, and Evaluation Metrics}
\label{what_was_run}
\centering
\begin{tabular}{|l|p{3cm}|p{3cm}|l|}
\hline
\textbf{Dataset} & \textbf{Encoders} & \textbf{Classifiers} & \textbf{Metrics}\\
\hline
SMS Spam & Bloom filter, Raw & Logistic Regression & $F_1$ (positive)\\ \hline
EKG & Bloom filter, Raw & XGBoost & $F_1$ (positive)\\ \hline
Adult 50K & Bloom filter, Raw, PCA, LDA & XGBoost, DNN & AUC \\\hline
CDC Diabetes & Bloom filter, Raw, PCA, LDA & XGBoost, DNN & AUC\\ \hline
MNIST & Bloom filter, Raw & DNN, CNN & Macro $F_1$ \\ \hline
Fashion MNIST & Bloom filter, Raw & DNN, CNN & Macro $F_1$\\ \hline
\end{tabular}
\end{table}
\section{Results}
\subsection{Accuracy}
\begin{figure}[t]
    \vspace{-2mm}
    \centering
    \includegraphics[width=.95\textwidth]{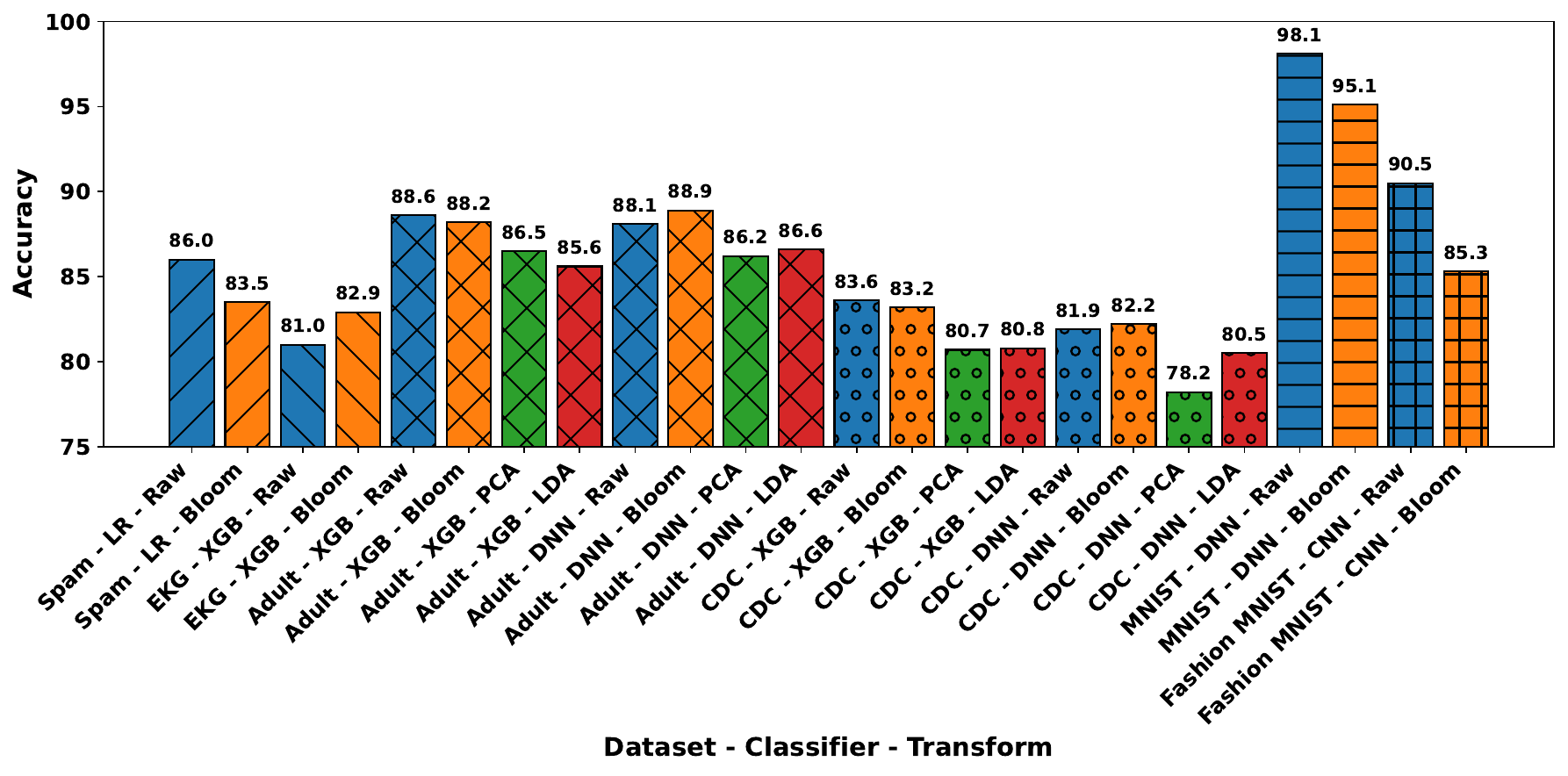}
    \caption{Accuracy by dataset, classifier, and transform (color = transform, hatch = dataset).}
    \label{fig:accuracy_comparison}
    \vspace{-6mm}
\end{figure}

The accuracy of the classifiers across datasets and transforms is shown in Fig. \ref{fig:accuracy_comparison}. Overall, the Bloom filter transform achieves performance that is comparable to raw data across several datasets, with modest improvements in some cases. In all comparisons, models are trained using the same hyperparameters, with only the input representation differing. The rationale for using identical hyperparameters across representations is to isolate the effect of the input encoding.

For PCA and LDA, the number of components was selected such that the resulting representation size was approximately matched to the Bloom filter encoding, enabling a fair comparison of predictive performance under similar memory constraints.

For example, on the EKG dataset, Bloom filter encoding improves accuracy from 81.0\% (raw) to 82.9\% (+1.9\%), and for the Adult 50K dataset using a DNN classifier, accuracy increases from 88.1\% (raw) to 88.9\% (+0.8\%). In cases where raw data performs slightly better, such as Adult 50K with XGBoost (88.6\% raw vs.\ 88.2\% Bloom, -0.4\%), the differences are small.

\begin{table}[t]
\centering
\caption{Adult 50K performance across 5-fold cross-validation (mean AUC $\pm$ standard deviation).}
\label{tab:adult_variance}
\begin{tabular}{|l|c|c|}
\hline
\textbf{Representation} & \textbf{XGBoost} & \textbf{DNN} \\
\hline
Raw   & 88.76 $\pm$ 0.37 & 88.13 $\pm$ 0.62 \\
\hline
Bloom & 88.21 $\pm$ 0.13 & 88.66 $\pm$ 0.34 \\
\hline
PCA   & 86.73 $\pm$ 0.43 & 86.58 $\pm$ 0.62 \\
\hline
LDA   & 86.07 $\pm$ 0.55 & 86.22 $\pm$ 0.53 \\
\hline
\end{tabular}
\end{table}

\begin{table}[t]
\centering
\caption{EKG200 performance across 5-fold cross-validation (mean accuracy $\pm$ standard deviation).}
\label{tab:ecg_variance}
\begin{tabular}{|l|c|c|}
\hline
\textbf{Representation} & \textbf{XGBoost} & \textbf{DNN} \\
\hline
Raw   & 81.02 $\pm$ 1.18 & 80.11 $\pm$ 2.36 \\
\hline
Bloom & 82.93 $\pm$ 1.05 & 83.21 $\pm$ 1.57 \\
\hline
PCA   & 79.77 $\pm$ 1.42 & 78.64 $\pm$ 2.41 \\
\hline
LDA   & 77.86 $\pm$ 1.64 & 76.92 $\pm$ 2.78 \\
\hline
\end{tabular}
\end{table}

To assess the stability of the results, Table~\ref{tab:adult_variance} reports mean and standard deviation across cross-validation folds for the Adult dataset. The relatively small standard deviations indicate that performance is consistent across folds and not driven by a particular data split. Notably, Bloom filter representations exhibit variance that is comparable to, and in several cases lower than, that of raw features, indicating that the encoding introduces no additional instability despite the use of hashing.

A similar analysis for the ECG200 dataset is shown in Table~\ref{tab:ecg_variance}. In this case, Bloom filter encoding not only achieves higher predictive performance but also maintains comparable or lower variance across folds. The observed improvements exceed the magnitude of the standard deviation, indicating that the gains are robust and not attributable to random variation.

Compared to dimensionality reduction methods such as PCA and LDA, Bloom filter encodings generally achieve comparable accuracy across the tabular datasets considered. For example, on Adult 50K (XGB), Bloom achieves 88.2\% accuracy compared to 86.5\% (PCA) and 85.6\% (LDA), and on CDC Diabetes (DNN), Bloom achieves 82.2\% versus 78.2\% (PCA) and 80.5\% (LDA). These results suggest that Bloom filter representations retain useful predictive structure while providing a compact encoding.

For image datasets, raw data achieves higher accuracy, with Bloom filter encodings showing moderate degradation. On MNIST, accuracy decreases from 98.1\% (raw) to 95.1\% (Bloom), a 3.0 percentage point reduction, and on Fashion MNIST from 90.5\% to 85.3\% (-5.2\%). These results are consistent with the loss of spatial structure introduced by hashing, which can limit the effectiveness of models that rely on local feature relationships.

\subsection{Compression}
\begin{figure}[tb]
    \vspace{-2mm}
    \centering
    \includegraphics[width=.95\textwidth]{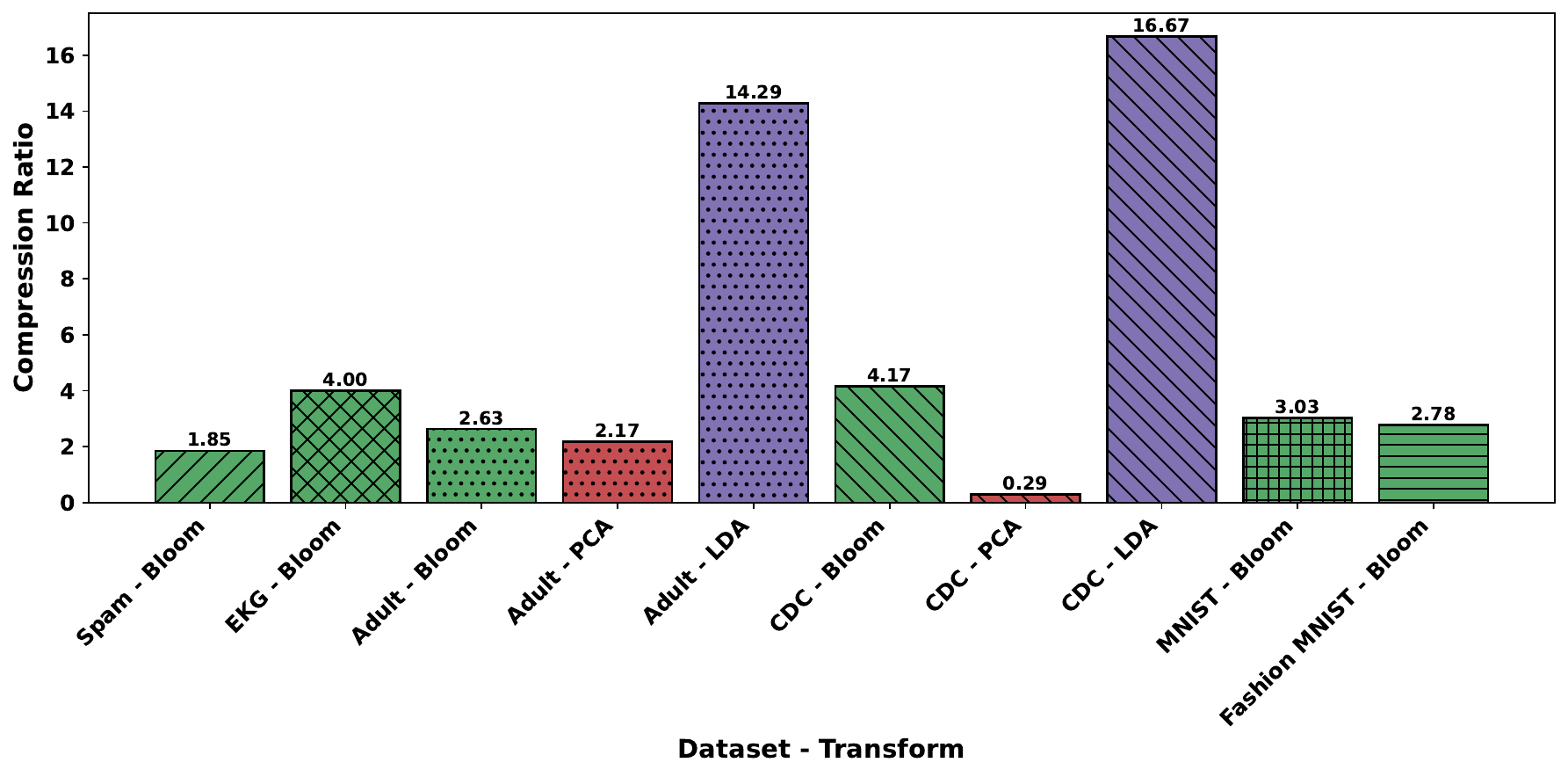}
    \caption{Compression ratio by dataset and transform (color = transform, hatch = dataset).}
    \label{fig:compression_comparison}
    \vspace{-6mm}
\end{figure}

Compression results are shown in Fig. \ref{fig:compression_comparison}. Bloom filter encodings provide consistent reductions in representation size relative to raw data. For example, on the EKG dataset, the Bloom filter representation achieves a 4.00$\times$ reduction, meaning each sample is approximately 25\% of the size of the raw representation.

Linear transforms such as LDA achieve higher compression ratios (e.g., 14.29$\times$ on Adult 50K and 16.67$\times$ on CDC Diabetes), but this is accompanied by a larger reduction in predictive performance. In contrast, Bloom filter encodings provide moderate compression while maintaining higher accuracy across these datasets.

Compared to PCA, Bloom filter encodings often achieve both higher compression and improved predictive performance (e.g., 2.63$\times$ vs.\ 2.17$\times$ compression on Adult 50K, and 4.17$\times$ vs.\ 0.29$\times$ on CDC). These results highlight a trade-off: Bloom filters provide a balanced compromise between compression and predictive performance, rather than maximizing compression alone.

\subsection{Entropy and Bit Occupancy}
\begin{figure}[t]
    \vspace{-2mm}
    \centering
    \includegraphics[width=.95\textwidth]{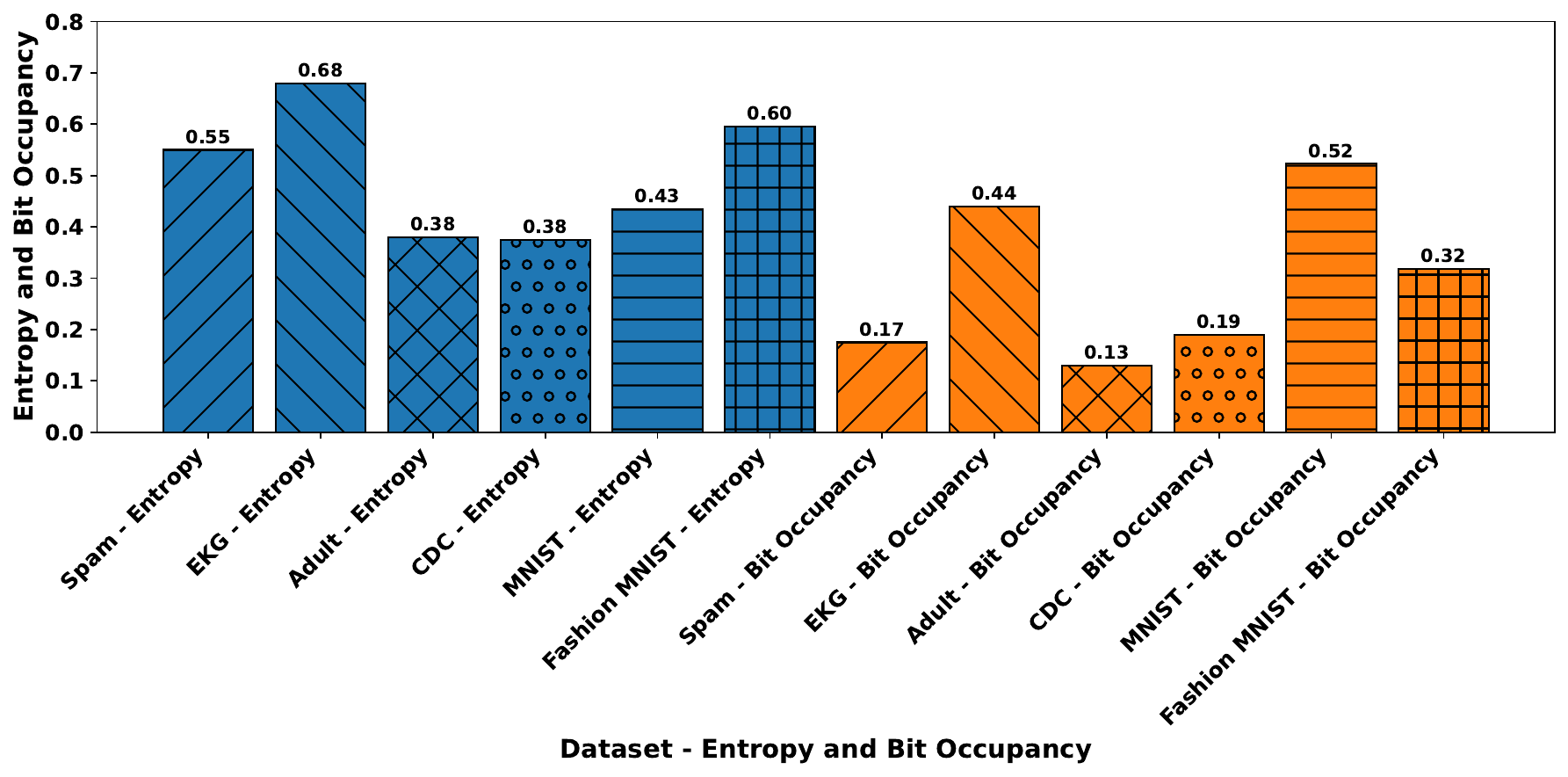}
    \caption{Entropy and bit occupancy per dataset for Bloom filter encodings (color = metric, hatch = dataset).}
    \label{fig:privacy_comparison}
    \vspace{2mm}
\end{figure}

Figure \ref{fig:privacy_comparison} summarizes entropy and bit occupancy, computed using Equations \ref{eq:entropy} and \ref{eq:bit_occupancy}. Entropy values range between 0.38 and 0.68 across datasets, indicating moderate uncertainty per bit in the encoded representations. Bit occupancy ranges from 0.13 to 0.60, showing that the Bloom filters are neither overly sparse nor fully saturated.

These results suggest that the chosen Bloom filter parameters produce representations with a balance between information density and collision effects. Higher occupancy values approach the region where entropy is maximized, while lower occupancy reflects sparser encodings with fewer collisions. While entropy and occupancy provide useful indicators of representation characteristics, they do not constitute formal privacy guarantees such as differential privacy. Instead, they reflect the degree to which feature information is distributed and obscured within the encoded space.

\subsection{Bloom Filter Configuration Sweep}
\begin{figure}[!t]
    \vspace{-2mm}
    \centering
    \includegraphics[width=.775\textwidth]{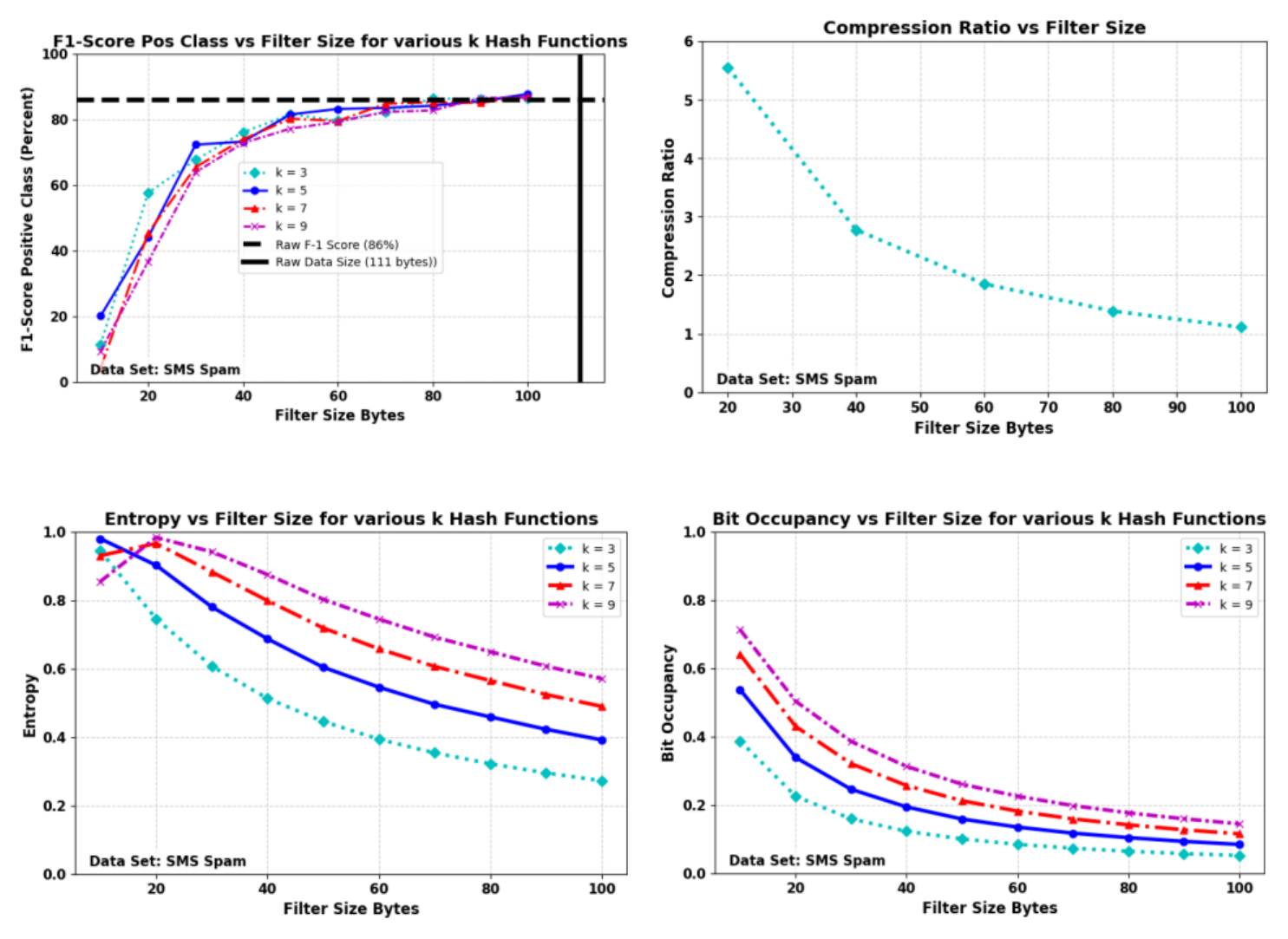}
    \caption{SMS Spam Dataset -- Sweeping Bloom filter size and number of hash functions.}
    \label{fig:sms_all}
    \vspace{-6mm}
\end{figure}

To analyze parameter sensitivity, we varied the Bloom filter size $m$ and number of hash functions $k$ on the SMS dataset (Fig. \ref{fig:sms_all}). Predictive performance generally improves as the filter size increases, reflecting reduced collision effects and improved representational capacity. Compression follows the opposite trend: smaller filters provide higher compression but typically reduce predictive performance.

The interaction between $m$ and $k$ also shows a trade-off. Moderate values of $k$ generally perform well, while larger values can increase collisions when the filter size is fixed. 

These results illustrate that Bloom filter parameters can be tuned to balance predictive performance, compression, and representation characteristics. Identifying optimal parameter settings for specific applications remains an area for future work.
\FloatBarrier
\section{Conclusion}
In this paper, we presented a Bloom filter--based preprocessing approach for machine learning. Models trained on encoded Bloom filter representations achieve performance comparable to raw data across several datasets, and in some cases outperform standard dimensionality reduction methods such as PCA and LDA under comparable representation sizes. Bloom filter encodings also provide consistent memory savings while introducing a degree of data obfuscation through hashing and collisions.

The results demonstrate that Bloom filter representations retain sufficient structure for effective learning while offering a compact, flexible encoding. Rather than explicitly preserving distances, the transform maintains approximate similarity relationships that are adequate for many classification tasks. This approximation is more effective for structured and tabular data, while performance can degrade for data with strong spatial dependencies such as images.

Future work includes extending this approach to regression and distributed learning settings, where compression and communication efficiency are critical. We also plan to evaluate additional datasets and model architectures, and to investigate techniques such as puncturing and folding to improve compression and representation efficiency. Hybrid approaches combining Bloom filter encodings with selected structured features may further improve performance while maintaining compact representations. 

Overall, Bloom filter encodings provide a practical preprocessing option that enables trade-offs between predictive performance, representation size, and data obfuscation. While the method does not provide formal privacy guarantees and requires parameter tuning (e.g., $m$, $k$), it offers a flexible and effective representation for a range of machine learning tasks.

\end{document}